\documentclass[11pt]{article}

\usepackage{appendix}
\usepackage{tikz}
\usepackage{amssymb,amsmath,amsfonts,dsfont}
\usepackage{graphicx}
\usepackage{subfig}
\usepackage[figuresright]{rotating}
\usepackage[applemac]{inputenc}
\usepackage{graphicx}
\usepackage{dcolumn}
\usepackage{bm}
\usepackage{amscd,amsthm}
\usepackage{ifthen}
\usepackage{booktabs}
\usepackage{bm}
\usepackage{makecell} 

\usepackage{mdframed}
\usepackage{listings} 
\usepackage{xcolor} 
\usepackage{tcolorbox} 

\usepackage[skip=0pt]{caption}

\usepackage{hyperref}
\hypersetup{
	bookmarks=true,         
	unicode=false,          
	pdftoolbar=true,        
	pdfmenubar=true,        
	pdffitwindow=false,     
	pdfstartview={FitH},    
	pdfauthor={Kun Wang},     
	pdfsubject={Subject},   
	pdfcreator={Kun Wang},   
	pdfproducer={Producer}, 
	pdfnewwindow=true,      
	colorlinks=true,       
	linkcolor=red,          
	citecolor=green,        
	filecolor=magenta,      
	urlcolor=cyan           
}
\usepackage{breakurl}

\usepackage[
backend=bibtex,
url=false,isbn=false,doi=false,
date=year,
hyperref=auto,
style=alphabetic,
natbib=true,
sorting=nty,
firstinits=true,
maxnames=2, maxbibnames=10,
]{biblatex}
\AtEveryBibitem{%
  \clearname{translator}%
  \clearlist{publisher}%
  \clearfield{pagetotal}%
  \clearname{editor}
  \clearfield{pages}
  \clearfield{series}
  \clearlist{address}
  \clearfield{address}
  \clearname{address}
  \clearname{volume}
  \clearfield{volume}
  \clearname{number}
  \clearfield{number}  
} 

\usepackage{tikz}
\usepackage{mathtools}
\usepackage{pgfplots}
\usepgfplotslibrary{groupplots} 
\usetikzlibrary{pgfplots.groupplots}
\usetikzlibrary{matrix,arrows,decorations.pathmorphing}
\usepackage{pgfplots,pgfplotstable}
\usepgfplotslibrary{fillbetween}
\usetikzlibrary{shadows,arrows}

\usepackage{changepage} 

\usetikzlibrary {arrows.meta} 
\usetikzlibrary{fit} 
\usetikzlibrary{backgrounds}

\usepgfplotslibrary{colorbrewer}
\pgfplotsset{compat=newest}
\usetikzlibrary{external}

\usetikzlibrary{calc}
\usetikzlibrary{positioning}
\usetikzlibrary{
  positioning,
  arrows.meta,
  shapes.geometric,
  shapes.symbols,
  shapes.callouts,
  fit,
  backgrounds,
  calc
}

\usepackage{graphics}
\usepackage{comment}
\usepackage{readarray} 
\usepackage{caption} 
\usepackage{xcolor}         
\usepackage{footnote}
\makesavenoteenv{tabular}
\usepackage{refcount}

\usepackage{hyperref}       
\usepackage{url}            
\usepackage{amsfonts}       
\usepackage{nicefrac}       
\usepackage{microtype}      

\usepackage{afterpage}

\newdimen\nodeDist
\usepackage{url}
\usepackage{breakurl}
\usepackage{outlines} 
\usepackage{changepage} 

\usepackage{rays_defs_18}

\definecolor{darkgreen}{RGB}{50, 168, 82}

\newtcolorbox{promptbox}[1][]{
  colback=gray!5!white,      
  colframe=gray!75!black,    
  title=\textbf{Prompt Template}, 
  fonttitle=\bfseries, 
  boxrule=0.5pt,    
  fontupper=\ttfamily,  
  width=0.9\linewidth,    
  center,    
  #1
}

\begin{document}

\begin{center}
	
	{\bf{\LARGE{
        Test-time RL alignment exposes task familiarity artifacts in LLM benchmarks
	}}}
	
	\vspace*{.2in}
	
	{\large{
			\begin{tabular}{cccc}
			Kun Wang
            ~~~~~~ Reinhard Heckel \\
            School of Computation, Information and Technology, Technical University of Munich \\
            Munich Center for Machine Learning
			\end{tabular}
	}}
	
	\vspace*{.05in}
	
	\begin{tabular}{c}


	\end{tabular}
	
	\vspace*{.1in}
 
	\today
	
	\vspace*{.1in}
	
\end{center}

\begin{abstract}
Direct evaluation of LLMs on benchmarks can be misleading because comparatively strong performance may reflect task familiarity rather than capability. 
The train-before-test approach controls for task familiarity by giving each model task-relevant training before evaluation, originally through supervised finetuning. However, suitable training data is often hard to come by, and evaluation results vary with the data chosen.
In this paper, we propose a two-stage test-time reinforcement learning (RL) alignment method for train-before-test. 
First, RL with a single sample provides a first alignment of the model to the task format, and second, test-time RL with majority-voting reward aligns the model to the benchmark distribution.
Our test-time RL alignment method aligns similarly well as SFT-based train-before test, but without requiring a task-specific training set.
On a domain-specific benchmark without training data, we show that direct evaluation underestimates base models which perform substantially better once aligned, yielding a more faithful evaluation of their capabilities. 
Moreover, for reasoning tasks, the performance gap between fine-tuned models and their base models largely disappears after alignment, suggesting that many gains from RLVR/SFT reported in the literature are not a  difference in reasoning capability, but rather artifacts of task familiarity.

\end{abstract}
{
\begin{figure*}[h]
    \centering
    \def\imgwidth{15cm}
   \includegraphics[width=\imgwidth]{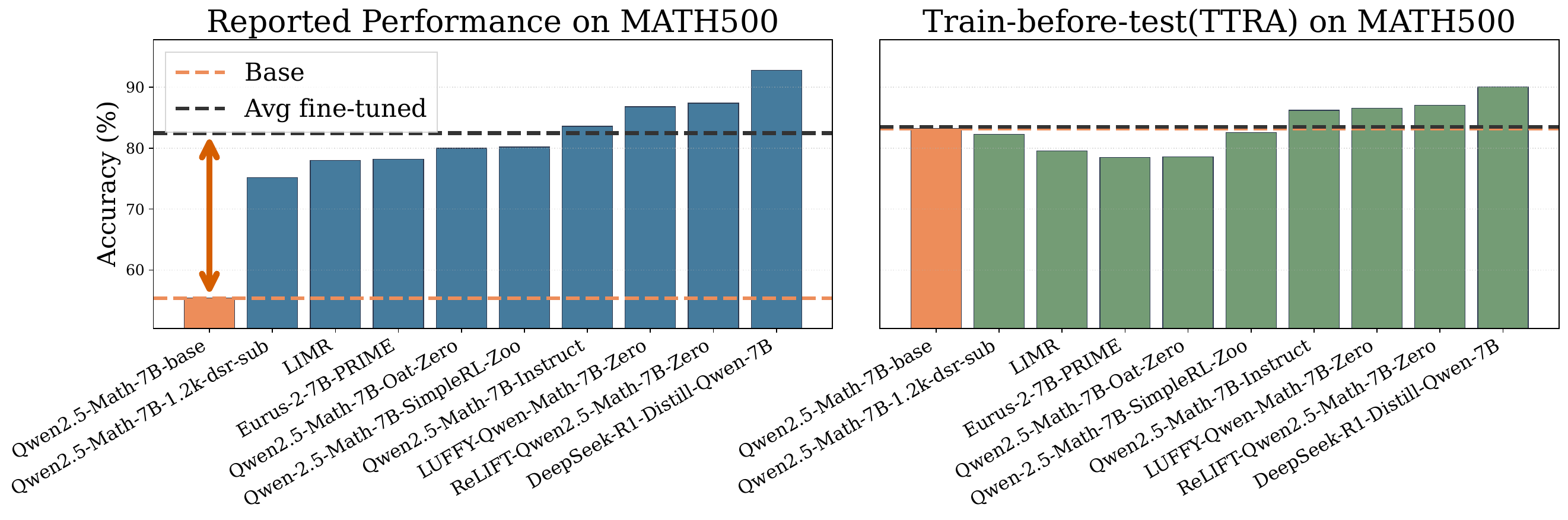}
    \caption{
    Fine-tuning gains are largely diminished after task alignment. \textbf{Left:} The reported performance reveals substantial accuracy gaps between base models and their fine-tuned variants. \textbf{Right:} After applying TTRA to all models, the base model's performance increases significantly, nearly matching that of the fine-tuned models, suggesting that many gains from RLVR/SFT reported in the literature are not a  difference in reasoning capability, but rather artifacts of task familiarity. 
    }
    \label{fig:reasoning_qwen_7b}
\end{figure*}
}
\section{Introduction}
\label{sec:intro}
Large Language Models (LLMs) have demonstrated remarkable performance across a diverse range of complex tasks. 
To measure improvements in model, training, and data, robust evaluation methodologies are critically important. 
Typically, performance is assessed on benchmarks such as MMLU for broad knowledge~\cite{hendrycks2020measuring}, GSM8K~\cite{cobbe2021gsm8k} and MATH~\cite{lightman2023lets} for mathematical reasoning, \cite{hager2024evaluation} for clinical decision making, and \cite{dominguez2024lawma} for legal classification tasks, to give some examples.
However, direct benchmark evaluation can systematically underestimate a model's underlying capabilities, as performance on a benchmark depends on a model's familiarity with the task beyond its intrinsic ability, as noted by~\citet{dominguez2024training}.

This evaluation challenge is especially relevant for reasoning models such as OpenAI's o1~\citep{jaech2024openai} and DeepSeek-R1~\citep{guo2025deepseek}, which have attracted significant attention. 
Several works report substantial reasoning performance gains from post-training LLMs with  
supervised fine-tuning (SFT) and reinforcement learning with verifiable rewards (RLVR)~\cite{wu2025generalization,liu2025understanding}.

However, recent work questions these findings. 
\citet{yue2025does}, for example, argues that RLVR may not genuinely incentivize reasoning ability. 
\citet{hochlehnert2025sober} questions  current progress in RLVR, finding that for models already fine-tuned on math reasoning tasks (e.g., DeepSeek-R1-Distill), further RLVR yields no meaningful improvement, whereas base models benefit significantly from RLVR. However, base models are often unfamiliar with the task, which may confound such comparisons. 

To enable fair comparisons of models on benchmarks, \citet{dominguez2024training} proposed the \textit{train-before-test} approach, in which a model is given task-relevant training. 
\citet{dominguez2024training,zhang2025train} demonstrate that supervised fine-tuning (SFT) on a benchmark-specific training set before evaluation harmonizes model rankings and removes apparent performance gaps between newer and older models. 

However, a challenge of such SFT-based train-before-test is that it can be difficult to obtain a suitable (in-distribution) training set for a given benchmarks. 
For example, \citet{dominguez2024training} used the validation sets of the respective benchmarks as training sets, but sometimes they are quite different from the benchmark: for example for MMLU, a knowledge multiple-choice benchmark, the validation set consists largely of reading-comprehension examples. 
For reasoning tasks, it is challenging to identify an appropriate SFT training dataset, as different choices of training data can yield considerably different performance.

In this paper, we propose a \textbf{dataset-free test-time reinforcement learning alignment (TTRA) method for train-before-test}.
Our method updates the model, but only on the benchmark's tasks, with minimal supervision. 
It consists of a one-shot reinforcement learning step based on a single labeled example to provide the model with the necessary initial signal for task format alignment, and a second reinforcement learning step based on majority-voting reward on the benchmark to generalize and approach the model's underlying performance. 

Our findings are as follows:

\textbf{First}, we study TTRA in a similar setup as SFT-based train-before test~\cite{dominguez2024training} to validate it. We find that it harmonizes model rankings across related benchmarks similar to SFT-based train-before-test, but with the added advantage of not using training data. We also demonstrate that it is stable on reasoning benchmarks and verify the consistency of the method on hold-out data. 

\textbf{Second}, we consider reasoning and demonstrate that
 TTRA reduces the performance differences between base models and their fine-tuned reasoning variants substantially (see Figure~\ref{fig:reasoning_qwen_7b}).
This suggests that the observed performance gains from many RLVR and SFT methods are not advances in reasoning ability but an artifact of the evaluation being confounded by the task familiarity acquired during fine-tuning.

\textbf{Third}, we consider a domain-specific benchmark for clinical decision making, and find that direct evaluation substantially underestimates the performance of base models. 
Compared with direct evaluation, TTRA unlocks the model's performance and demonstrates that while scores are initially low due to task unfamiliarity, the model achieves accuracy comparable to human doctors after alignment, without requiring a separate training dataset.

\section{Test-time reinforcement learning alignment (TTRA) for train-before-test}
\label{sec:oattrl_method}
\begin{figure*}[t]
  \centering
  \includegraphics[width=0.8\linewidth]{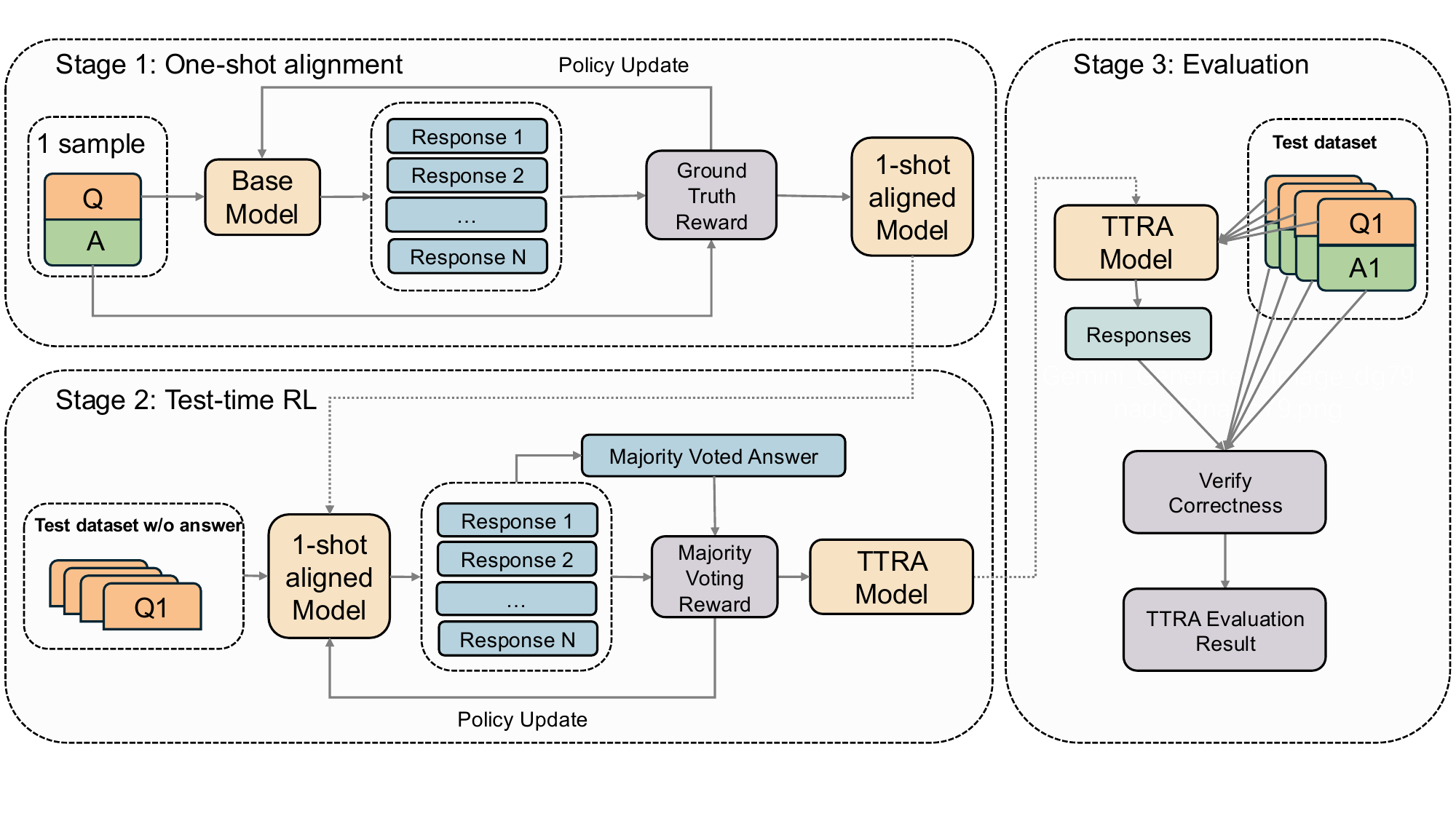}
  \caption{Overview of the TTRA base train-before-test evaluation pipeline.}
  \label{fig:oattrl_pipeline}
\end{figure*}

In this section we motivate and introduce our training-set free train-before-test alignment method for improved model evaluation.

\subsection{Motivation}
Most commonly, an LLM is directly evaluated on a benchmark. 
However, as observed by~\citet{dominguez2024training} this can yield misleading rankings, as superior performance of one model may reflect greater exposure to the test task rather than truly better performance.
\textit{Train-before-test}~\cite{dominguez2024training} addresses this by aligning models before testing them through training. 
\citet{zhang2025train} demonstrated that SFT-based train-before-test provides a more consistent measure of model performance. 

However, SFT-based train-before-test faces  limitations that stem from its dependency on the training dataset. 
For reasoning tasks, the dependence on the training dataset introduces substantial instability in evaluation. Applying supervised fine-tuning to the same base model with different reasoning-oriented training datasets can yield substantially different performance, making the resulting model comparisons
highly sensitive to the choice of training data rather than the model's relative reasoning abilities.

Moreover, curating suitable validation sets is difficult for many benchmarks. 
For many widely used reasoning benchmarks, such as AIME, 
no official training or validation set exists, forcing researchers to curate datasets from other sources. 
Even an official training set is provided, it may not be in-distribution. An example is MMLU~\citep{hendrycks2020measuring}, where the provided training data focuses on reading comprehension, while the test set evaluates general knowledge across various domains. 
Also, for domain-specific benchmarks, such as clinical decision-making~\citep{hager2024evaluation}, building a dataset for supervised train-before-test would require expensive annotations from domain experts.

\subsection{Test-time RL alignment (TTRA)} 
To overcome the limitations of SFT-based train-before-test, we introduce a dataset-free train-before-test method. 
Our approach leverages insights from recent self-improvement methods that require no ground-truth labels and have demonstrated that LLMs possess an inherent potential for test-time adaptation. For instance, prior work has explored entropy minimization as a training signal~\citep{cui2025entropy, agarwal2025unreasonable, gao2025one}. Furthermore, recent studies have shown that test-time training can be effective for LLMs. \citet{zuo2025ttrl} uses majority voting to generate pseudo-labels for test-time reinforcement learning (TTRL), while \citet{h2025learningjobtesttimecurricula} proposes test-time curriculum RL to improve performance.

However, unsupervised TTRL can fail if a model lacks a minimal level of task competence, as it may not generate meaningful responses from which to derive a useful reward signal. Conversely, one-shot reinforcement learning~\citep{wang2025reinforcement} can provide a crucial initial alignment signal but often causes the model to overfit to the limited example, failing to generalize across the entire test set and thus capping its potential performance. Therefore, our approach combines those two steps. 

By combining these two approaches, the one-shot step provides a first alignment of the model to the task, 
and the test-time RL with majority-voting reward step aligns the model to the benchmark distribution. 
This leads to an test-time train-before-test method referred as test-time RL alignment (TTRA) and illustrated in Figure~\ref{fig:oattrl_pipeline}. In this work, we use Group Relative Policy Optimization (GRPO)~\citep{shao2024deepseekmath} as the policy optimization algorithm.

\textbf{Stage 1: One-shot task format alignment}
In the initial stage, we randomly select a single labeled sample from the available data to perform one-shot reinforcement learning. The model learns the fundamental structure and format expectations of the task through this minimal supervision.

\textbf{Stage 2: Test-time reinforcement learning}
After the initial alignment, the model possesses basic task understanding, enabling effective test-time adaptation. We employ TTRL using majority voting as the reward signal. This allows the model to further refine its performance across the entire test distribution. The second stage ensures comprehensive alignment with the test dataset while leveraging the model's own generations as learning signals, eliminating the need for additional labeled data. 

Our method maintains the benefits of train-before-test while eliminating the dependency on a separate training dataset, requiring only a single labeled sample that is typically obtainable from benchmark.
\section{Validation of TTRA-based train-before-test}
\label{sec:validating_oattrl}

In this section, we validate the effectiveness of TTRA. First, we study TTRA in a similar setup as the original train-before-test works~\cite{dominguez2024training,zhang2025train} considered. We demonstrate that TTRA harmonizes model rankings across related benchmarks, achieving results similar to SFT-based train-before-test, but with the added advantage of not using training data. Furthermore, we show that our method exhibits better stability compared to SFT-based train-before-test on a reasoning benchmark. Finally, we verify the method's consistency on held-out test sets to address potential concerns regarding overfitting to the test set.

\subsection{TTRA can harmonize LLM rankings}
\label{sec:harmonize}
\begin{figure}[t]
    \centering
    
    \begin{minipage}[t]{0.48\textwidth}
        \centering
        \includegraphics[width=\textwidth]{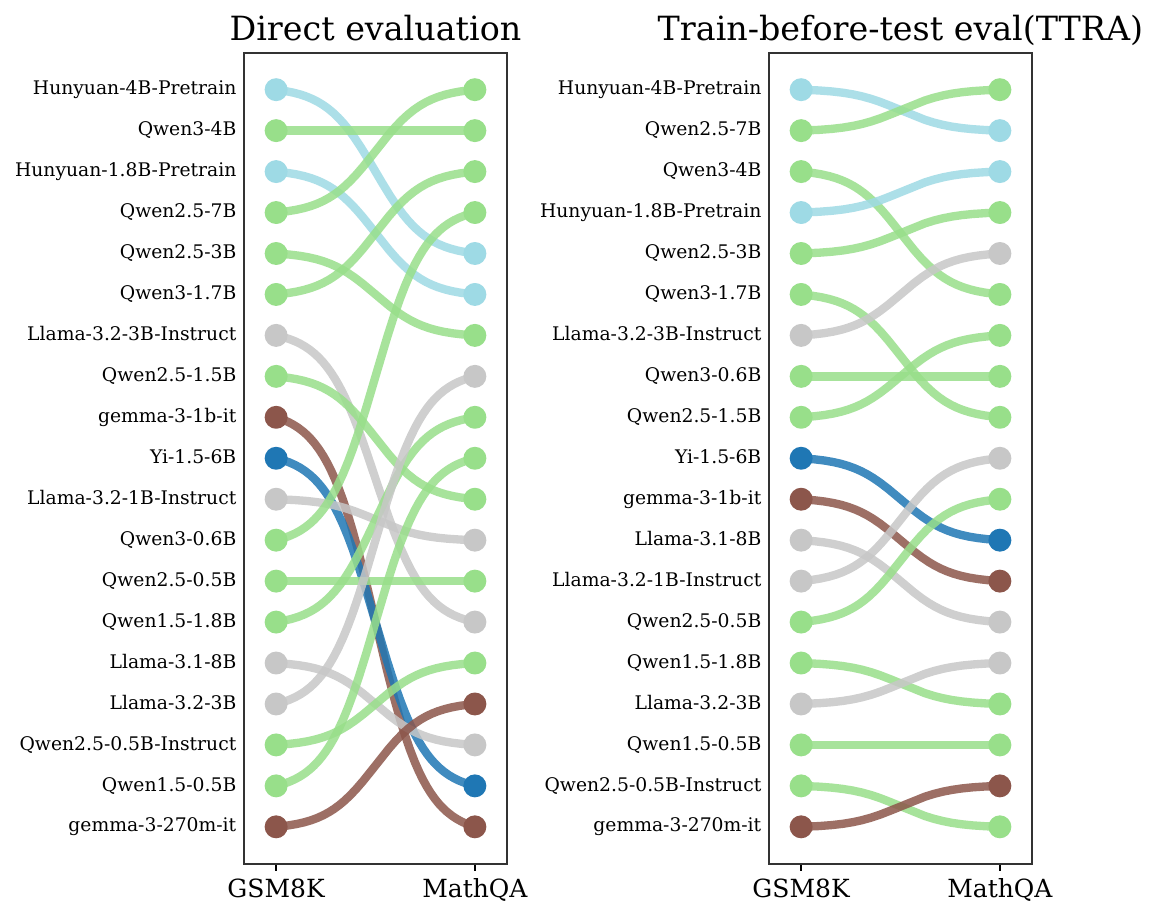}
        \caption{
        TTRA harmonizes model rankings between GSM8K and MathQA. 
        \textbf{Left:} Direct evaluation shows highly discordant model rankings across the two benchmarks. 
        \textbf{Right:} After TTRA alignment, the rankings become substantially more consistent and harmonized.
        }
        \label{fig:gsm8k_mathqa_comparison}
    \end{minipage}
    \hfill
    \begin{minipage}[t]{0.48\textwidth}
        \centering
        \includegraphics[width=\textwidth]{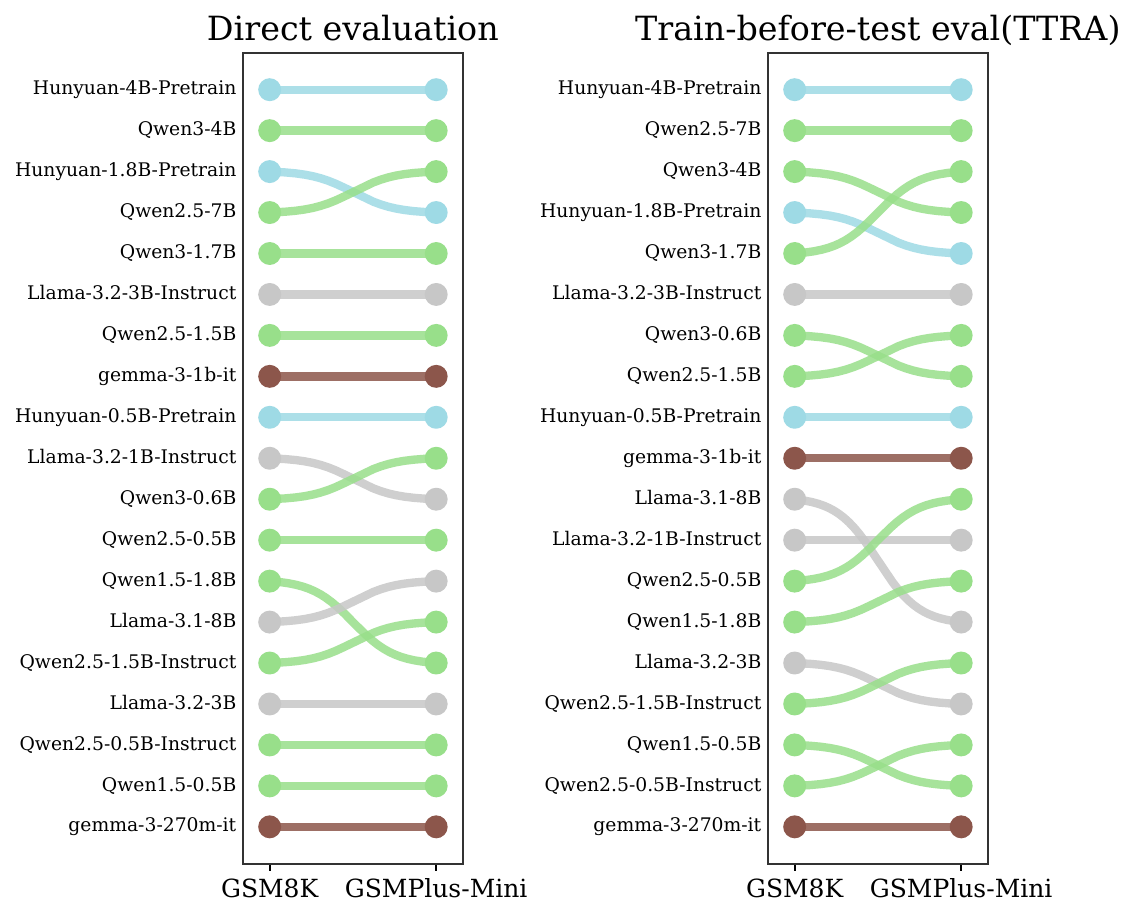}
        \caption{
        TTRA preserves existing ranking harmony between GSM8K and GSM-plus-mini. 
        \textbf{Left:} Direct evaluation shows already consistent rankings for these similar-format tasks. 
        \textbf{Right:} TTRA evaluation maintains this high consistency.
        }
        \label{fig:gsm8k_gsmplus_comparison}
    \end{minipage}
    
\end{figure}
One would expect that relative comparisons among models are relatively consistent for different benchmarks that target a given domain. 
However,~\citet{zhang2025train} revealed that model rankings often vary substantially across related tasks, and  demonstrated that SFT-based train-before-test can harmonize these inconsistent rankings. 

We demonstrate that our proposed TTRA framework achieves similar harmonization as SFT-based train-before-test, but without requiring SFT training data. 
We consider open-weight LLMs including models from Qwen \citep{qwen,qwen2.5,qwen2,qwen3technicalreport}, Hunyuan~\citep{hunyuan_collection_2025}, Yi~\citep{ai2024yi}, Gemma~\citep{gemmateam2025gemma3technicalreport}, and Llama \citep{grattafiori2024llama3herdmodels}. We evaluated those models on three mathematical benchmarks: GSM8K~\citep{cobbe2021gsm8k}, MathQA~\citep{amini-etal-2019-mathqa}, and GSM-plus-mini~\citep{li2024gsm}, with and without TTRA. For TTRA alignment, a single sample was randomly selected from the benchmark for one-shot training, and the remaining data was used for the TTRL stage and final evaluation.
Full details of the evaluated models and the corresponding training configurations are in Appendix~\ref{app:harmonize}.

As Figure~\ref{fig:gsm8k_mathqa_comparison}(left) shows, under direct evaluation, model rankings between GSM8K and MathQA are inconsistent. As  Figure~\ref{fig:gsm8k_mathqa_comparison}(right) shows, after TTRA train-before-test, the rankings become substantially more harmonized, with models maintaining more consistent ordinal positions across both benchmarks. 

Another expected property of a train-to-test method is that, if model rankings are already consistent, TTRA preserves this property. 
Figure~\ref{fig:gsm8k_gsmplus_comparison} demonstrate that TTRA indeed preserves the already-consistent rankings for models evaluated on the GSM8K and GSM-plus-mini benchmarks. 

\subsection{Stability analysis of TTRA and SFT-based train-before-test}
\label{sec:stability_analysis}
The reliability of SFT-based  train-before-test for task alignment depends on the choice of training dataset, and  evaluation scores can fluctuate significantly based on the alignment dataset. 

In this section, we compare the stability of TTRA and SFT-based train-before-test under controlled sources of randomness. 
For TTRA, we used three different random seeds to select the one-shot alignment sample, utilizing the remaining data for both TTRL and evaluation.
To quantify sensitivity of SFT based train-before-test, we fine-tuned the Qwen2.5-Math-1.5B model using a fixed budget of 1,000 samples. We randomly subsampled from 
AM-DeepSeek-Distilled~\citep{tian2025deepdistillenhancingllmreasoning}, Eurus-2-SFT~\citep{cui2025process,yuan2024implicitprm}, and OpenThoughts-3~\citep{guha2025openthoughtsdatarecipesreasoning}, whereas the S1K-1.1 dataset~\citep{muennighoff2025s1} is fully used as it contains exactly 1k samples.
Full training details are provided in Appendix~\ref{app:stability_analysis}.

\begin{figure}[t]
    \centering
    
    \begin{minipage}[t]{0.48\textwidth}
        \centering
        \def\imgwidth{7.5cm}
        \includegraphics[width=\imgwidth]{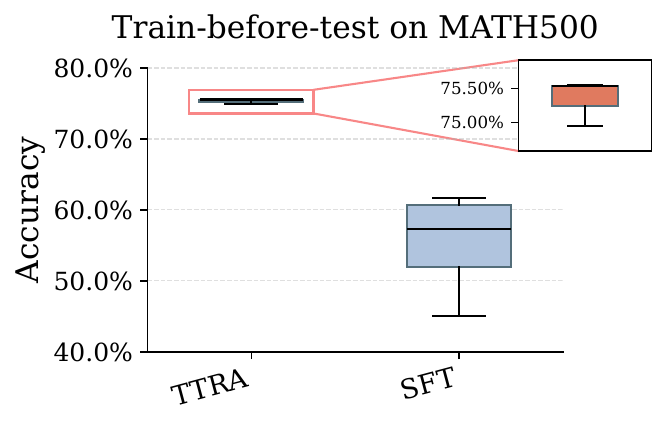}
        \caption{
Stability analysis on MATH500. 
Our dataset-free TTRA method is robust to random choices of one-shot sample selection ($\Delta<1\%$). 
In contrast, SFT-based train-before-test is highly sensitive to the training data used. 
        }
        \label{fig:stability_analysis}
    \end{minipage}
    \hfill
    \begin{minipage}[t]{0.48\textwidth}
        \centering
        \def\imgwidth{4.5cm}
        \includegraphics[width=\imgwidth]{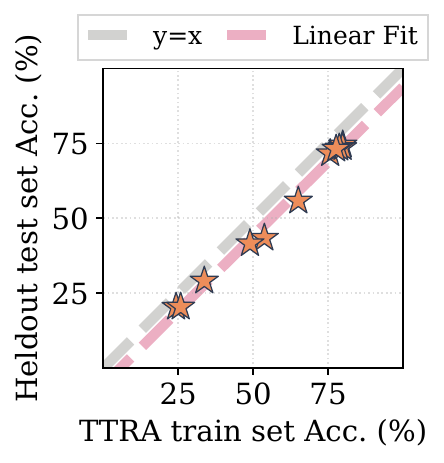}
        
        \caption{
Consistency check via split-half validation on MATH500. 
We compare accuracy on a strictly held-out test subset (y-axis) with accuracy on the subset used for TTRA optimization (x-axis). Each orange star corresponds to a model. The grey dashed diagonal ($y = x$) marks the ideal case where test and optimization performance are equal. The pink dashed line shows a linear fit across models. 
Performance is highly consistent across the two subsets, with the fitted line lying only slightly below the $y = x$ diagonal.
        }
        \label{fig:reasoning_heldout}
    \end{minipage}
    
\end{figure}

As illustrated in Figure~\ref{fig:stability_analysis}, TTRA is is robust, maintaining consistent performance across different random choices of the single example used for the one-shot step. 

In contrast, SFT-based train-before-test is quite sensitive to the data source used for training, as somewhat expected. On MATH500, accuracy fluctuates by over 16\% depending on the dataset used for SFT-based train-before-test. Extended experiments showing similar variation for SFT on AIME and OlympiadBench are in Appendix~\ref{app:stability_analysis}. This confirms that SFT-based train-before-test is sensitive to the training data, whereas TTRA avoids this issue. 
   
\subsection{Consistency study under a separate test set}
\label{sec:consistency_check}

The TTRL-alignment step only utilizes the question of the benchmark data, but not their labels. 
In principle, the observed performance gains of our method could therefore arise from robust task alignment (as desired) but also from potential overfitting to the benchmark, despite the fact that TTRA does not use test labels during training.
To study whether this occurs, we performed a split-half validation by randomly partitioning the test-data into two equally-sized subsets. The first half was used to update the model via TTRA, while the second served as a strictly held-out test set for evaluation.

As shown in Figure~\ref{fig:reasoning_heldout}, evaluations of models on the train and test split are essentially perfectly correlated and are very close to the $y=x$ diagonal. 
This experiment supports the validity of TTRA for train-before-test for evaluation and mitigates the overfitting concern. 

\section{TTRA train-before-test on reasoning models}
\label{sec:exp_reasoning}

{
\begin{figure*}
    \centering
    \def\imgwidth{15cm}
   \includegraphics[width=\imgwidth]{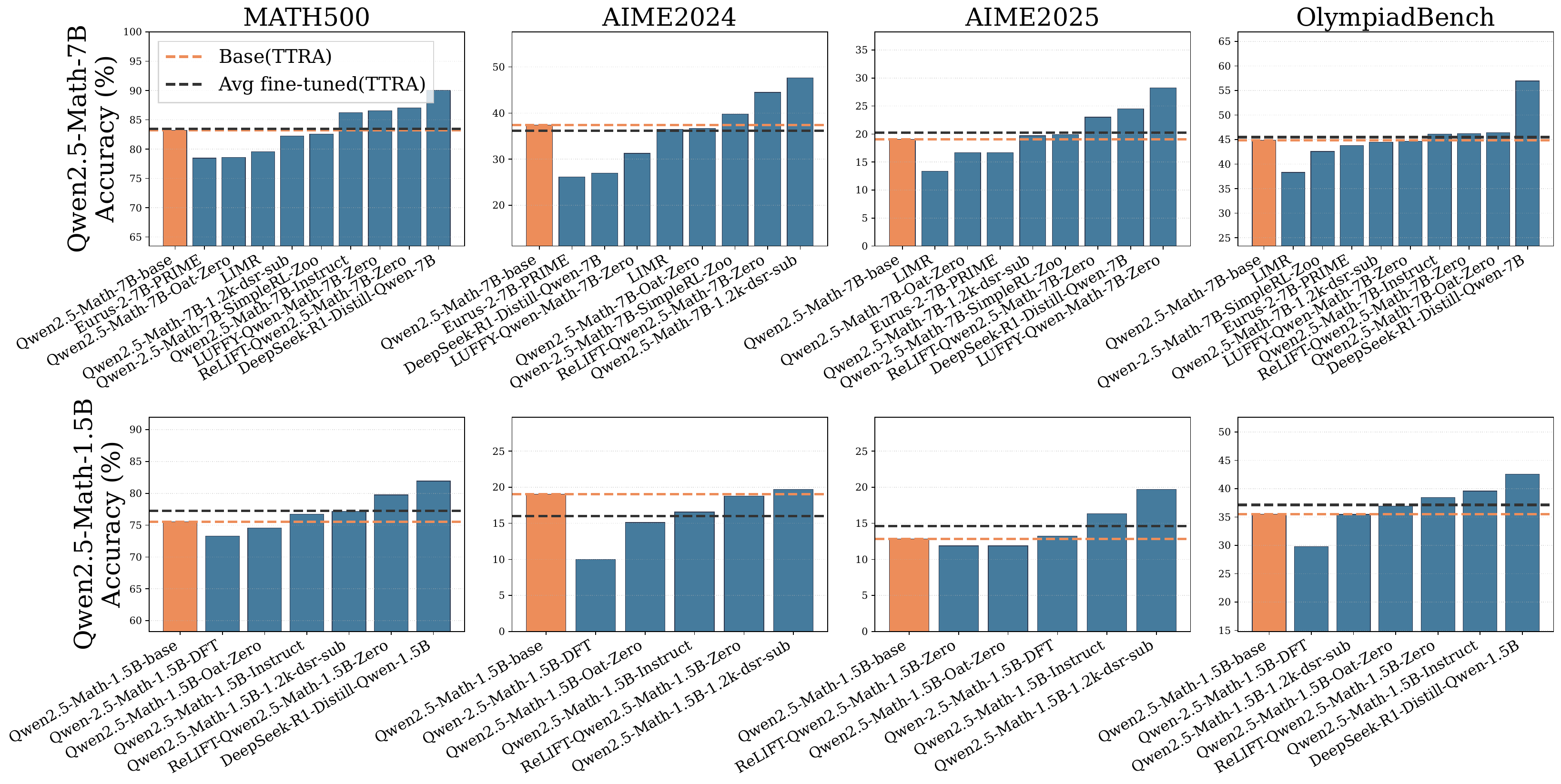}
    \caption{
    Performance comparison of Qwen2.5-Math base and fine-tuned models under TTRA evaluation. Results are shown for 7B models (top) and 1.5B models (bottom) across four benchmarks. The orange bars denote the base models, while blue bars denote various fine-tuned variants. The orange dashed line marks the base model's performance level, highlighting that the performance advantage of fine-tuned models largely disappears after applying TTRA.
    }
    \label{fig:reasoning_all}
\end{figure*}
}
Recently, several LLM fine-tuning recipes (including method and data) have been proposed, many of which report substantial performance gains on reasoning benchmarks. It is important to understand whether these recipes improve a model's reasoning capability.

Some prior works have studied whether RLVR enhances reasoning capabilities. 
For example, \citet{yue2025does} argues that the gains observed from RLVR may not reflect stronger underlying reasoning, but instead stem from improved sampling behavior, and that RLVR does not introduce new reasoning patterns.
Similarly, \citet{hochlehnert2025sober} finds that for models already fine-tuned on math reasoning tasks, such as DeepSeek-R1-Distill, additional RLVR yields little benefit, while base models show substantial improvements.
However, this comparison may be confounded by differences in task familiarity, as base models may perform poorly due to limited exposure under reasoning task rather than insufficient reasoning capability.


To address this gap, we evaluate several base models and their fine-tuned variants on math reasoning benchmarks. We apply TTRA to all models prior to evaluation to reduce disparities in task alignment, helping make comparisons less sensitive to task familiarity and more reflective of underlying reasoning capability.

\subsection{Experimental Setup}
We evaluated Qwen2.5-Math-1.5B/7B base models~\citep{yang2024qwen25mathtechnicalreportmathematical} and a range of  variants fine-tuned for reasoning tasks.
We include variants derived from Qwen2.5-Math-Instruct~\citep{yang2024qwen25mathtechnicalreportmathematical}, ReLIFT~\citep{ma2025learning}, Oat-Zero~\citep{liu2025understanding}, DeepSeek-R1-Distill~\citep{guo2025deepseek}, as well as Qwen2.5-Math-1.2k-dsr-sub, a model trained with a 1.2k DeepScalerR subset~\citep{wang2025reinforcement,deepscaler2025}.
At the 1.5B scale, we additionally include a model fine-tuned using the dynamic fine-tuning (DFT) method~\citep{wu2025generalization}. 
At the 7B scale, we further incorporate several reasoning variants, including PRIME~\citep{cui2025process,yuan2024implicitprm}, SimpleRL-Zoo~\citep{zeng2025simplerl}, LIMR~\citep{li2025limr}, and LUFFY~\citep{luffy}.


We evaluate these models on mathematical reasoning benchmarks, including MATH500~\citep{lightman2023lets}, AIME 2024 \& 2025, and OlympiadBench~\citep{he2024olympiadbench}.
For MATH500, we randomly select a sample from MATH500 with a difficulty level of 1 for the 1-shot alignment step.
For AIME, we perform the initial 1-shot alignment step on a randomly chosen AIME2023 question and the second TTRL-alignment on the combination of AIME2024\&2025 since the AIME dataset only contains 30 questions each.
Further training details and hyperparameters are in Appendix~\ref{app:details_exp_reasoning}.

\subsection{Many reported fine-tuning gains are largely diminished after TTRA}

The results for Qwen2.5-Math-7B are shown in Figure~\ref{fig:reasoning_qwen_7b}. Figure~\ref{fig:reasoning_qwen_7b} (left) presents the performance reported in the original model papers, where the scores for the base model and Qwen2.5-Math-7B-Instruct are taken from \citep{liu2025understanding}.

For direct evaluation, the fine-tuned variants appear to be significantly better than base models, with an average performance gain of $\Delta = 27.07\%$.
However, after applying TTRA to control for task familiarity the performance gap virtually vanishes (see Figure~\ref{fig:reasoning_qwen_7b} (right)). 
This suggests that the base model possesses largely similar reasoning abilities to the fine-tuned versions.

In Figure~\ref{fig:reasoning_all}, we extend this evaluation to two model scales and four reasoning benchmarks.
Consistent with the findings in Figure~\ref{fig:reasoning_qwen_7b}, the average performance gap between base models and fine-tuned variants is small under train-before-test with TTRA. 

Remarkably, the aligned base model not only matches the average but even outperforms several  fine-tuned variants.
However, we observe models like DeepSeek-R1-Distill, which maintain a consistent performance advantage even after TTRA alignment. This suggests that distillation from strong teacher models can improve reasoning capabilities that persist beyond task alignment.
We also evaluated additional Llama models and observed a similar pattern: performance gains diminish substantially after train-before-test. The results are reported in Appendix~\ref{app:details_exp_reasoning}, Table~\ref{tab:appendix_reasoning_comparison}.

Although fine-tuning methods can yield substantial performance gains, our results indicate that these improvements may be confounded by task familiarity. Consequently, some of the observed performance improvements may be an artifact of the evaluation, driven by the heightened task familiarity acquired during fine-tuning rather than by intrinsic advances in reasoning.

\section{LLMs can be underestimated on domain-specific benchmarks}
\label{sec:exp_cdm}

Many important real-world applications of LLMs arise in domain-specific settings, where task definitions and input formats can differ substantially from those in commonly used benchmarks. The community has  developed a growing number of domain-specific benchmarks. However, evaluating models directly on such benchmarks can systematically underestimate their true capabilities, because the model is unfamiliar with the task.

Clinical decision-making is a representative example. The MIMIC-CDM benchmark \cite{hager2024evaluation} evaluates models on complex clinical cases in which the model must infer an underlying pathology or diagnosis from structured evidence, including patient history, physical examination findings, laboratory test results, and imaging reports. 

This benchmark has two settings: a full-information setting, where all available clinical information is presented in a single prompt and the model directly predicts the diagnosis, and a conversational setting, where only the patient history is initially revealed and the model must iteratively request additional information (e.g., examination findings, labs, imaging) before making a diagnostic decision. 

We hypothesize that when LLMs are evaluated directly under these domain-specific benchmarks without any task alignment, their performance is systematically underestimated, reflecting unfamiliarity with the benchmark rather than a true lack of clinical decision-making ability. 

SFT-based train-before-test is not easily applicable to the MIMIC-CDM benchmark, because it does not provide an validation- or training set, and constructing one is expensive, as it requires expert annotations from licensed clinicians. 
In this section, we use TTRA to evaluate several base models on the MIMIC-CDM benchmark.

\subsection{Experimental setup}

We consider the MIMIC-CDM full-information(MIMIC-CDM-FI) setup of the benchmark, where we evaluate the models by providing all available clinical information in a single prompt. To manage context length effectively, we use the Qwen3 tokenizer for token counting. For specific sections, such as imaging reports, that exceed 1,000 tokens in length, we utilize GPT5-nano~\citep{singh2025openaigpt5card} to generate a concise summary of the content before feeding it into the model. All prompts used in this evaluation are detailed in Appendix~\ref{app:exp_cdm_details}.

\subsection{Task alignment can unlock LLMs' performance}
{
\begin{figure}[h]
    \centering
    \def\imgwidth{10cm}
   \includegraphics[width=\imgwidth]{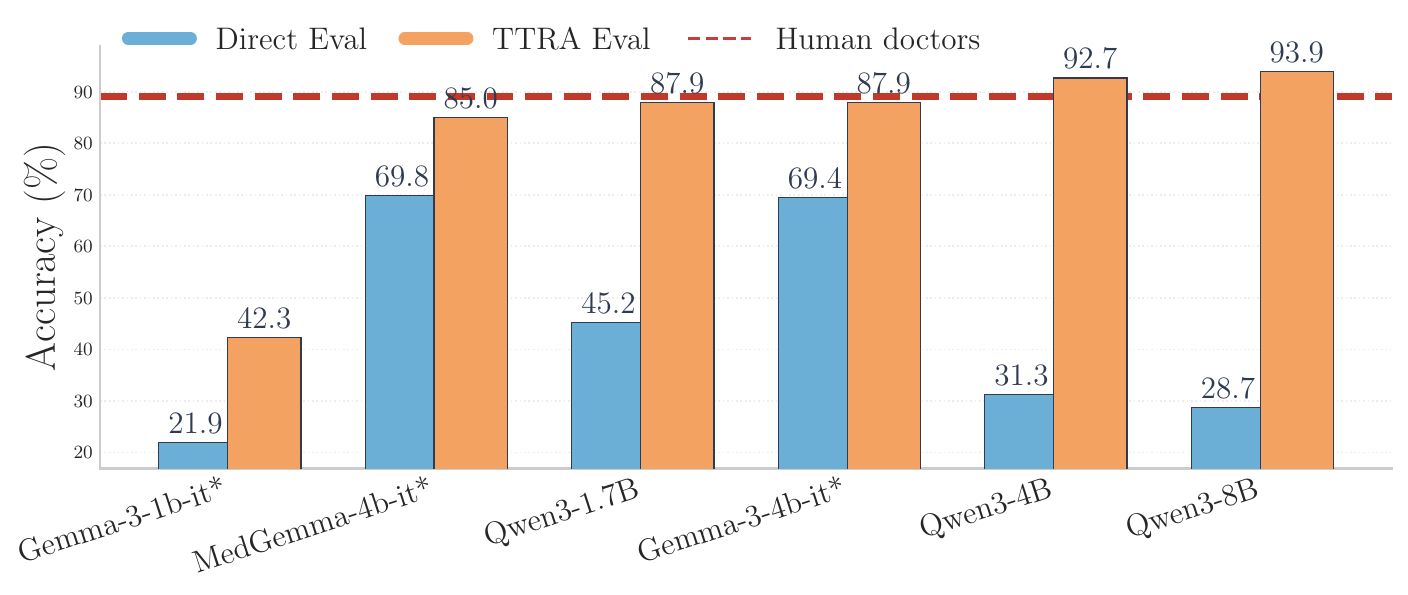}
    \caption{
Evaluation of models aligned via TTRA compared to human doctors and a directly evaluated baseline. 
For models marked with (*), base model performance is taken from \cite{hager2024evaluation}.
After applying TTRA, the models significantly outperform direct evaluation, indicating that the base models possess latent capabilities previously underestimated, and relative comparisons between models change. 
}
    \label{fig:cdm_qwen_7b}
\end{figure}
}

Figure~\ref{fig:cdm_qwen_7b} shows the clinical decision-making accuracy of several base models before and after alignment with TTRA, alongside a human doctor baseline reported in~\citet{hager2024evaluation}. Without alignment, the Gemma base models~\citep{gemmateam2025gemma3technicalreport} perform poorly. For instance, Gemma-3-1B reaches only 21.9\% accuracy, which is worse than random guessing.

After applying TTRA, all Gemma models improve substantially. Gemma-3-1B increases from 21.9\% to 42.3\%, while MedGemma-3-4B-it rises from 69.4\% to 86\%. 
A similar pattern is observed across the Qwen3 family, where TTRA consistently leads to large gains over direct evaluation. 
Notably, the aligned Qwen3-4B and 8B models achieve performance comparable to the human doctor baseline. 
These results indicate that train-before-test alignment can substantially alter the apparent performance of base models in clinical decision-making tasks, revealing capabilities that are underestimated under direct evaluation.

These findings suggest that performance gaps of models for domain-specific benchmarks must not be due to a lack of knowledge, but can be due to limited familiarity with the task format. TTRA effectively bridges this gap, unlocking capabilities already present in models, and can therefore help evaluate LLMs on domain-specific benchmarks more faithfully.

\section{Discussion and limitations}
\label{sec:discussion}
In this section, we discuss the stage-wise effectiveness of TTRA, compare the compute costs of different evaluation methods, and discuss limitations of our method.

\subsection{Stage-wise ablation study of TTRA}
\label{sec:ablation_study}
To validate the necessity of the two steps in our two-step framework, we conducted an ablation study to decouple the contributions of 1-shot alignment and TTRL. 
We find, as noted before, the two steps to be complementary: 1-shot alignment establishes the initial task alignment needed to reliably start the optimization, while TTRL enables continued adaptation to the benchmark distribution. For some models, removing 1-shot alignment is expected to cause TTRL to fail, because the majority-vote signal no longer provides meaningful rewards. Conversely, relying only on 1-shot alignment caps performance prevents the model from reaching its full potential.

To analyze the contribution of both stages in TTRA, we compare the two-step pipeline against the single-step variants. 
We evaluate on two distinct setups: Qwen2.5-Math-1.5B on the MATH500 benchmark, and the general-purpose Qwen2.5-1.5B on the MIMIC-CDM-FI benchmark.

\begin{figure}
  \vskip 0.2in
  \begin{center}
    \centerline{\includegraphics[width=0.98\columnwidth]{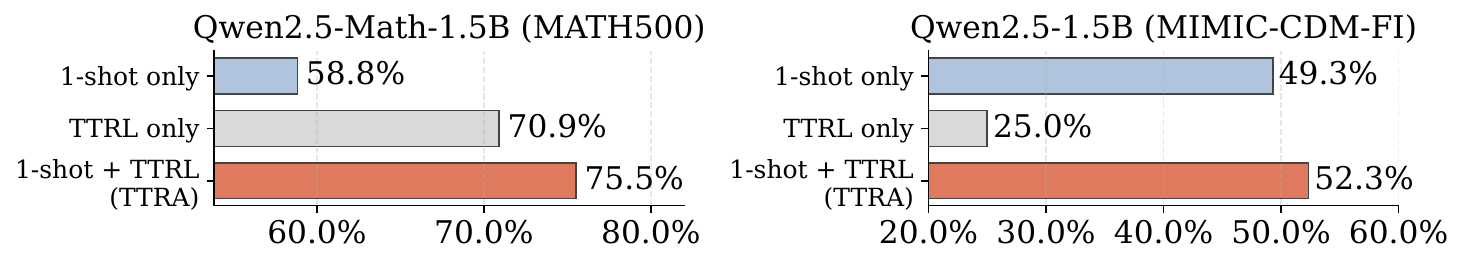}}
    \caption{
Stage-wise ablation study of TTRA. \textbf{Left:} For Qwen2.5-Math-1.5B on the MATH500 benchmark, both 1-shot alignment and TTRL contribute positively, with the hybrid TTRA approach achieving the highest accuracy. \textbf{Right:} For Qwen2.5-1.5B on the MIMIC-CDM-FI benchmark, removing 1-shot alignment causes TTRL to fail, highlighting the critical role of initial alignment in unfamiliar domains.
}
    \label{fig:ablation_study}
  \end{center}
\end{figure}

The results are shown in Figure~\ref{fig:ablation_study}. On MATH500, where the Qwen2.5-Math-1.5B already possesses strong reasoning capabilities, both 1-shot alignment and TTRL improve performance. The TTRL-only approach yields more gains over 1-shot only, and combining them in TTRA achieves the highest performance. 

The results on MIMIC-CDM-FI reveal that  while 1-shot alignment effectively adapts the model to the complex clinical format, the TTRL-only approach fails, yielding an accuracy of only 25.0\% due to severe reward hacking. We observe that without the initial alignment provided by the 1-shot example, the TTRL optimization collapses to a degenerate solution predicting a single pathology (specifically appendicitis) in 99.93\% of the responses. This confirms that 1-shot alignment is necessary for effective TTRL for some benchmarks.

\subsection{Compuational cost}
\label{sec:computational_cost}
In this section, we examine the computational cost of three evaluation methods: direct evaluation, train-before-test with SFT, and train-before-test with TTRA. Of course those costs depend on the size of the training set used by SFT, as well as the specifics of the model and benchmark. 

We took the Qwen3-0.6B model and evaluated it on the GSM8K benchmark. For direct evaluation, we sample responses with a temperature of 0.6, sampling 32 responses per question. For TTRA, we use a single randomly sampled example from the GSM8K training split for one-shot alignment and allocate the test split to the TTRL stage. For the SFT baseline, we combined MetaMath~\citep{yu2023metamath} and Orca-Math~\citep{mitra2024orca}, which are also used in original train-before-test work~\citep{dominguez2024training}. Detailed experimental settings are provided in the Appendix~\ref{app:discussion_details}.

The results are shown in Table~\ref{tab:time_analysis}. Direct evaluation incurs the lowest computational cost and is used as a reference. As discussed in Section~\ref{sec:harmonize}, both SFT- and TTRA-based train-before-test approaches can harmonize LLM rankings. However, the SFT-based method requires approximately 10 hours of training, resulting in a relative cost of about $32\times$ compared to direct evaluation. In contrast, TTRA achieves significantly lower overhead, requiring only a 6.6$\times$ relative cost.
Overall, these results suggest that, in the train-before-test setting, TTRA can achieve similar algnment performance to SFT with lower computational cost on GSM8K.


\begin{table}[h]
\centering
\caption{Evaluation time and relative computational cost of direct evaluation and train-before-test methods (SFT and TTRA) using Qwen3-0.6B on GSM8K.}
\label{tab:time_analysis}
\begin{small}
\begin{tabular}{lcc}
\toprule
\textbf{Method} & \textbf{Time}& \textbf{Relative time} \\
\midrule
Direct Evaluation & 19 min & 1.0 \\

\midrule
Train-before-test (SFT) &  & 32$\times$ \\
\hspace{1.5em} \textit{SFT} & 10 h & \\
\hspace{1.5em} \textit{Inference} & 19 min& \\

\midrule
Train-before-test (TTRA) &  & 6.6$\times$  \\
\hspace{1.5em} \textit{One-shot RL} & 15 min & \\
\hspace{1.5em} \textit{TTRL} & 92 min & \\
\hspace{1.5em} \textit{Inference} & 19 min & \\

\bottomrule
\end{tabular}
\end{small}
\end{table}

\subsection{Limitation}
The TTRA framework uses majority voting to compute the reward signal during the TTRL-alignment. This reward construction is not well-suited to tasks such as code generation, where correctness is not reliably captured by agreement among sampled outputs; thus, majority voting does not provide a dependable learning signal.

Furthermore, our evaluation on MIMIC-CDM is limited to the full-information setting. We have not yet explored the conversational setting, which would require agentic workflows for iterative diagnosis. Extending our approach to agentic workflows and corresponding benchmarks remains an interesting direction for future work.


\subsection*{Acknowledgements}
The authors gratefully acknowledge the computing time made available for this
paper by the LRZ and MCML, as well as on the high-performance computer at the NHR Center of TU Dresden.
Moreover, the authors are supported by the German Federal Ministry of Education and Research, and the Bavarian State Ministry for Science and the Arts. 

\printbibliography{}


\newpage
\appendix

\section{
Details for Section~\ref{sec:validating_oattrl}: TTRA validation and stability
}
\subsection{Implementation details and additional results for Section~\ref{sec:harmonize}}
\label{app:harmonize}
\textbf{Training details.}
Our implementation is based on the verl framework~\citep{sheng2024hybridflow}.
We use GRPO~\citep{shao2024deepseekmath} for training. All models are trained using the AdamW optimizer with a cosine learning rate scheduler and a linear warmup over 10\% of the total steps. During training, responses are sampled with a temperature of 0.6, using 32 rollouts and a maximum sequence length of 1024 tokens.
The batch size is set to 1 for 1-shot RL and 4 for TTRL alignment.
For GSM8K and GSM-plus-mini, models are trained with a learning rate of $1\mathrm{e}{-6}$ for 100 steps, followed by $5\mathrm{e}{-7}$ for 300 steps. For MathQA, models are trained with a learning rate of $1\mathrm{e}{-6}$ for 100 steps, followed by $5\mathrm{e}{-8}$ for 300 steps. Since MathQA is a multiple-choice dataset, we want to prevent the model from collapsing to predicting a single option. To this end, we rotate the answer choices and construct multiple rotated variants for the 1-shot training sample.
All models are trained on H100 GPUs.

For data selection, we randomly select one example from the GSM8K training split and use the test split for TTRL and evaluation. For both MathQA and GSM-plus-mini, we randomly select one example from the test set and use the remaining examples for TTRL and evaluation.

\textbf{Verification and evaluation.}
For verification, we use math-verify~\citep{Kydlicek_Math-Verify_Math_Verification} on GSM8K and GSM-plus-mini.
For MathQA, we use exact string matching on either the final boxed answer or, if no boxed answer is present, the first line of the model output. This accounts for models that directly output the answer without intermediate reasoning.
During evaluation, we generate 32 responses per prompt with a temperature of 0.6, a top-$k$ of 0.95, and a maximum length of 1024 tokens.

We use the following prompt for GSM8K and GSM-plus-mini:
\begin{promptbox}[title=Prompt for GSM8K and GSM-plus-mini]
Q: \{Question\} 

Let's think step by step and output the final answer within \textbackslash boxed\{\}.
\end{promptbox}

We use the following prompt for MathQA:

\begin{promptbox}[title=Prompt for MathQA]
Let's think step by step and output the final answer with only the correct option letter enclosed in boxed\{\}.
\\
\\
Question: \\ \{Question\} 
\\
\\
Options:

A. ...

...
\\
\\
Answer:
\end{promptbox}

We provide visual illustrations of the rank harmonization effect discussed in Section~\ref{sec:harmonize}. Figure~\ref{fig:gsm8k_mathqa_comparison} depicts the ranking dynamics between GSM8K and MathQA. Under direct evaluation, the rankings exhibit significant discordance with numerous crossing lines, indicating that format differences obscure true relative capabilities. However, applying TTRA effectively aligns the tasks, substantially harmonizing the rankings. 
Furthermore, Figure~\ref{fig:gsm8k_gsmmini_comparison_corr} demonstrates the non-disruptive nature of our method on the closely related GSM8K and GSM-plus-mini benchmarks. For these naturally aligned tasks, TTRA preserves the existing high consistency, confirming that our framework corrects format-induced discrepancies where they exist without negatively impacting performance or rankings on already concordant tasks.

{
\begin{figure}[h]
    \centering
    \def\imgwidth{8cm}
   \includegraphics[width=\imgwidth]{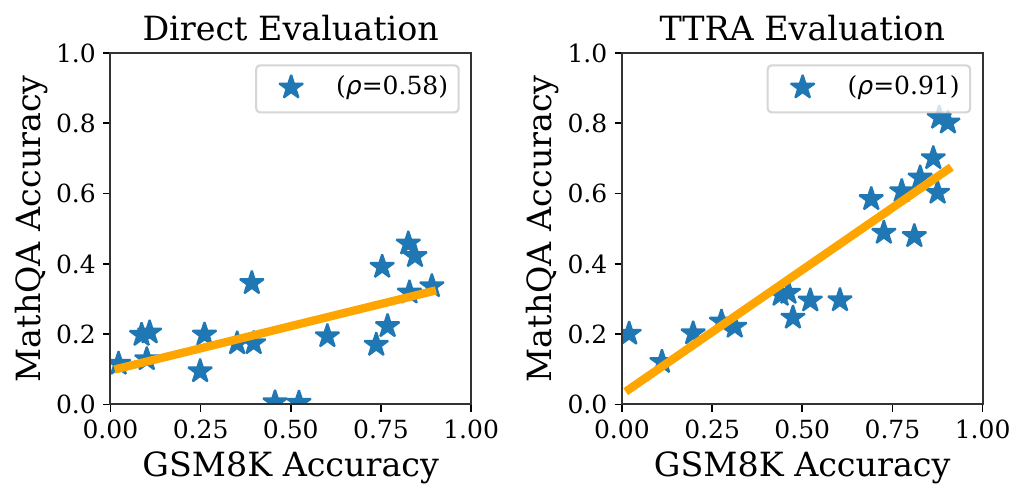}
    \caption{
TTRA increases the ranking correlation between GSM8K and MathQA.\textbf{Left:} Direct evaluation shows a moderate correlation ($\rho=0.58$)1. \textbf{Right:} After TTRA alignment, the correlation significantly increases ($\rho=0.91$).
}
    \label{fig:gsm8k_mathqa_comparison_corr}
\end{figure}
}
{
\begin{figure}[h]
    \centering
    \def\imgwidth{8cm}
   \includegraphics[width=\imgwidth]{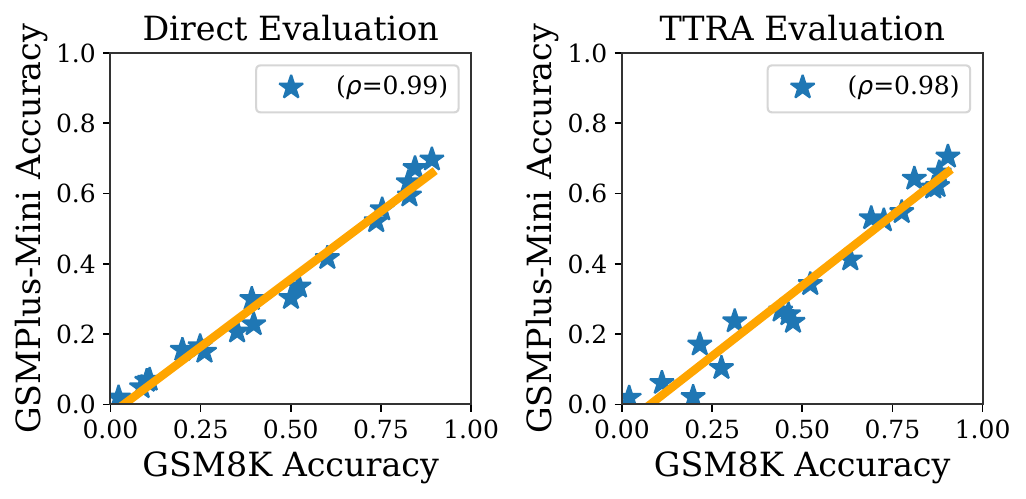}
    \caption{
TTRA maintains a high correlation for already harmonized tasks. \textbf{Left:} Direct evaluation of the GSM8K and GSM-plus-mini benchmarks shows a high correlation. \textbf{Right:} TTRA alignment preserves this high correlation.
}
    \label{fig:gsm8k_gsmmini_comparison_corr}
\end{figure}
}
\subsection{Implementation details and additional results for Section~\ref{sec:stability_analysis}}
\label{app:stability_analysis}

For SFT training, we follow the configuration in~\citet{muennighoff2025s1}. We train the models for 5 epochs with a learning rate of $1\text{e-}5$, a linear warmup over 5\% of the training steps, $\beta_1 = 0.9$, $\beta_2 = 0.95$, and a weight decay of $1\text{e-}4$. 
For TTRA training, we use the same setup as described in Appendix~\ref{app:details_exp_reasoning}, but with a different random seed for selecting the one-shot alignment sample.

We also provide some addition results of SFT sensitivity analysis discussed in Section~\ref{sec:stability_analysis}. Figure~\ref{fig:sft_performance} illustrates the performance of Qwen2.5-Math-1.5B fine-tuned on four distinct datasets (AM, s1k-1.1, Eurus-2, and OpenThoughts) across three benchmarks: MATH500, AIME2024, and OlympiadBench. We observe substantial performance fluctuations across all evaluated tasks, further confirms that SFT evaluation scores are heavily depends on the specific choice of training data.

{
\begin{figure}[h]
    \centering
    \def\imgwidth{12cm}
   \includegraphics[width=\imgwidth]{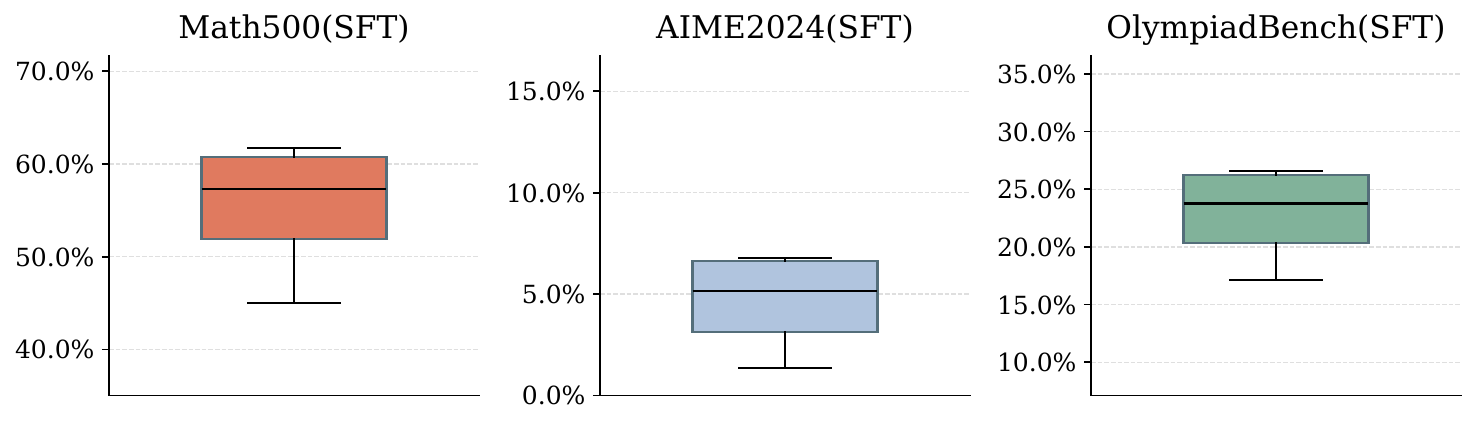}
    \caption{
Sensitivity of SFT to training data selection. We evaluate the performance of the base model fine-tuned on various open-source datasets with a fixed data budget (1k samples). The results reveal significant performance variance across different data sources.
}
    \label{fig:sft_performance}
\end{figure}
}

\subsection{Implementation details for Section~\ref{sec:consistency_check}}
For the consistency analysis, we sample 250 examples from the MATH500 benchmark for TTRA training and reserve the remaining 250 for held-out evaluation.
We follow the same training and evaluation setup as described in Appendix~\ref{app:details_exp_reasoning}, but use a smaller learning rate of $1\mathrm{e}{-6}$ during the TTRL alignment stage due to the smaller training set.

For this experiment, we evaluated Qwen2.5-Math-1.5B, along with its official Instruct variant~\citep{yang2024qwen25mathtechnicalreportmathematical} and several fine-tuned versions, including ReLIFT~\citep{ma2025learning}, Oat-Zero~\citep{liu2025understanding}, and a model trained on a 1.2k subset of DeepScalerR~\citep{wang2025reinforcement,deepscaler2025}. In addition, we included some other Qwen models: Qwen2.5-0.5B, Qwen2.5-1.5B~\citep{qwen2.5}, Qwen2-1.5B, as well as Qwen2-Math-1.5B~\citep{qwen2}. Finally, we also evaluated models from a different architecture family: Llama-3.2-1B-Instruct and Llama-3.2-3B-Instruct~\citep{grattafiori2024llama3herdmodels}.

\section{Details for Section~\ref{sec:exp_reasoning}: reasoning evaluation setup
}
\label{app:details_exp_reasoning}
In this section, we provide details of the experimental setup and additional results for the reasoning tasks discussed in Section~\ref{sec:exp_reasoning}. 

\paragraph{Training details} 
All models are trained with the AdamW optimizer, a cosine learning rate schedule, and a linear warmup over 10\% of the total training steps. For both one-shot alignment and TTRL, we use GRPO~\citep{shao2024deepseekmath} for reinforcement learning, sampling 32 responses per question at a temperature of 0.6.
We configured the 1-shot alignment stage with a learning rate of $5\text{e-}7$ for 100 epochs, generating 32 rollouts per step with a maximum length of 3072 tokens for MATH500 and AIME2024 \& 2025, and maximum length of 2500 tokens for the OlympiadBench. 
In the subsequent TTRL stage, we trained for 300 steps with a learning rate of $5\text{e-}6$ and a batch size of 4. We maintained a rollout size of 32, which was utilized for both optimization and majority voting. During the training, we used the math-verify~\citep{Kydlicek_Math-Verify_Math_Verification} for verifying the correctness and majority voting.

\paragraph{Verification and evaluation.}
We also use the math-verify for verification.
During the evaluation, we used the temperature $0.6$ and top-$k$ 0.95,, generate 32 responses with maximum length of 3072 tokens for generation, and report the average score.

Following is the prompt we used:
\begin{promptbox}[title=Prompt for Section~\ref{sec:exp_reasoning}]
<|im\_start|>system

You are a helpful assistant.<|im\_end|>

<|im\_start|>user

\{Question\} 

Let's think step by step and output the final answer within \textbackslash boxed\{\}. <|im\_end|>

<|im\_start|>assistant 
\end{promptbox}

\paragraph{Additional results}
In addition to the Qwen models, we also evaluate Llama models. We use the same user content as the prompt, but without the chat template. 
The results are summarized in Table~\ref{tab:appendix_reasoning_comparison}. Compared to direct evaluation, the performance gaps between base models and their fine-tuned variants are substantially reduced after applying TTRA, further supporting the conclusions drawn in Section~\ref{sec:exp_reasoning}.

\begin{table*}[h]
    \centering
    \small 
    \setlength{\tabcolsep}{15pt}
    
    \caption{Comparison between direct evaluation and evaluation using train-before-test with TTRA. Scores marked with * are reported from the original papers. Under direct evaluation, fine-tuned variants show large improvements; after applying TTRA, these gains are substantially smaller.}
    \label{tab:appendix_reasoning_comparison}
    
    \begin{tabular}{l c c c c c}
        \toprule
        & \multicolumn{2}{c}{\textbf{Math500}} & & \multicolumn{2}{c}{\textbf{OlympiadBench}} \\
        \cmidrule(lr){2-3} \cmidrule(lr){5-6}
        \textbf{Models} & \textbf{Direct} & \textbf{TTRA} & & \textbf{Direct} & \textbf{TTRA} \\
        \midrule
        
        \multicolumn{6}{c}{\textit{Llama-3.2-3B-NuminaQA}} \\
        \midrule
        Base         & $0.6^*$ & $40.52$ & & $0.1^*$ & 9.59 \\
        Oat-Zero-3B~\cite{liu2025understanding}     & $50.0^*$ & $46.27$ & & $14.7^*$& 12.53   \\
          & \textcolor{red}{+49.4} & \textcolor{blue}{+5.75} & & \textcolor{red}{+14.6} & \textcolor{blue}{+2.94} \\
                
        \bottomrule
    \end{tabular}
\end{table*}

\section{Implementation details for Section~\ref{sec:exp_cdm}: domain-specific benchmarks}
\label{app:exp_cdm_details}
For the MIMIC-CDM-FI dataset, we randomly sampled 1 case from each pathology for the 1-shot alignment, which is 4 samples in total. And the remaining for the evaluation.

\paragraph{Training details.} 
We follow the training setup described in Appendix~\ref{app:details_exp_reasoning}, with the following differences: we set the maximum input length to 4000 tokens and the maximum output length to 1024 tokens. For one-shot alignment, we use a learning rate of $1\mathrm{e}{-7}$ for 100 steps. For TTRL alignment, we use the same learning rate for 300 steps for Qwen3 models, and $5\mathrm{e}{-8}$ for Gemma models.

\textbf{Verification and evaluation.}
For verification, we extract the answer from the final boxed content and perform exact string matching. Consistent with our previous settings, we generate 32 responses per prompt using a temperature of 0.6, a top-$k$ of 0.95, and a maximum sequence length of 1024 tokens. For accuracy calculation, we first compute the accuracy for each pathology and then average the results across the 4 pathologies.

The following prompt is used for Qwen3 models. For Gemma models, we use the same prompt without the chat template.

\begin{promptbox}[title=Prompt for Section~\ref{sec:exp_cdm}]
<|im\_start|>system

You are a helpful assistant.<|im\_end|>

<|im\_start|>user

You are an expert Clinical Decision Support System. Your goal is to analyze patient data against a provided set of diagnostic criteria to identify the most likely pathology.

\{Diagnostic Criteria\} 

\{Patient History\} 

\{Physical Examination\} 

\{Laboratory Tests\} 

\{Imaging Reports\} 

Based on the above information, identify the most likely pathology affecting the patient.
Please explain your reasoning, and then provide the final answer with only the correct pathology enclosed in \textbackslash boxed\{\}.<|im\_end|>

<|im\_start|>assistant 
\end{promptbox}

The prompt used to summarize the long report is shown below:

\begin{promptbox}[title=Prompt for report summary]
You are a medical artificial intelligence assistant. Your goal is to effectively, efficiently and accurately reduce text without inventing information. You want to return verbatim observations that are abnormal and of interest to a possible diagnosis of the patient. Normal observations can be combined. Do not invent information. Use medical abbreviations when possible to save characters. Put the most important information first. Please summarize the following result: \\
\{Report\}\\
Summary: 
\end{promptbox}

\section{Additional details and results for Section~\ref{sec:discussion}: discussions and limitations}
\label{app:discussion_details}
\subsection{Implementation details for Section~\ref{sec:ablation_study}}

In Section~\ref{sec:ablation_study}, we conduct a stage-wise ablation study of TTRA. For training on MATH500, we adopt the same settings as described in Appendix~\ref{app:details_exp_reasoning}. For the MIMIC-CDM-FI benchmark, we follow a similar setup to that in Appendix~\ref{app:exp_cdm_details}. However, since the models exhibit weaker capability on this benchmark, we use a learning rate of $5\mathrm{e}{-6}$ for one-shot alignment and $5\mathrm{e}{-8}$ for TTRL alignment.

\subsection{Implementation details for Section~\ref{sec:computational_cost}}

For Section~\ref{sec:computational_cost}, all training and inference are conducted on NVIDIA H100 GPUs. The TTRA training and evaluation follow the same setup as described in Appendix~\ref{app:harmonize}. For SFT training, we adopt a training procedure similar to that of~\citet{dominguez2024training}. The model is trained using the AdamW optimizer with a cosine learning rate scheduler and a warmup ratio of 1\%, a learning rate of $2\mathrm{e}{-6}$, and a weight decay of 0.1. The batch size is set to 64. Training data consist of a combination of MetaMath~\citep{yu2023metamath} and Orca-Math~\citep{mitra2024orca}.

\end{document}